\pdfoutput=1

\documentclass[11pt]{article}
\usepackage{CJKutf8}
\usepackage[russian,english]{babel}
\usepackage{arabtex}

\usepackage{ACL2023}

\usepackage{times}
\usepackage{latexsym}
\usepackage{xcolor}
\usepackage{float}
\usepackage{colortbl}

\usepackage[T1]{fontenc}

\usepackage[utf8]{inputenc}
\usepackage{utf8}
\setcode{utf8}

\usepackage{microtype}

\usepackage{inconsolata}
\usepackage{xspace}
\usepackage{graphicx}
\usepackage{booktabs}
\usepackage{array,multirow}
\usepackage{float}
\usepackage{subcaption}

\usepackage[most]{tcolorbox}

\usepackage{todonotes}

\newcommand\verythinrule{\specialrule{.02em}{0.3em}{0.2em}}

\newcommand{\eg}{\textit{e.g.},\xspace}
\newcommand{\ie}{\textit{i.e.},\xspace}

\definecolor{ctemplate}{rgb}{0.23, 0.30, 0.45}
\definecolor{cword}{rgb}{0, 0, 0.7}

\newcommand{\template}[1]{\texttt{\textcolor{ctemplate}{``#1''\xspace}}}
\newcommand{\placeholder}[1]{\texttt{\textcolor{ctemplate}{#1\xspace}}}
\newcommand{\ttag}[1]{\texttt{\textcolor{cword}{\{#1\}\xspace}}}
\newcommand{\tfeat}[1]{\texttt{\textcolor{cword}{#1\xspace}}} 

\definecolor{cexample}{rgb}{0.23, 0.30, 0.45}
\newcommand{\exinline}[1]{\textcolor{cexample}{``#1''\xspace}}
\newcommand{\fillin}[1]{\textcolor{cexample}{#1\xspace}}

\newcommand{\en}{\textsc{en}\xspace}
\newcommand{\es}{\textsc{es}\xspace}
\newcommand{\It}{\textsc{it}\xspace} 
\newcommand{\fr}{\textsc{fr}\xspace}
\newcommand{\de}{\textsc{de}\xspace}
\newcommand{\sv}{\textsc{sv}\xspace}
\newcommand{\Fi}{\textsc{fi}\xspace}
\newcommand{\sk}{\textsc{sk}\xspace}
\newcommand{\ru}{\textsc{ru}\xspace}
\newcommand{\sw}{\textsc{sw}\xspace}
\newcommand{\zh}{\textsc{zh}\xspace}
\newcommand{\ar}{\textsc{ar}\xspace}

\newcommand{\cellbreaks}[2][c]{%
  \begin{tabular}[#1]{@{}l@{}}#2\end{tabular}}
  
\title{Empowering Cross-lingual Behavioral Testing\\of NLP Models with Typological Features}

\author{Ester Hlavnova \hspace{10pt} Sebastian Ruder \\
  Google Research \\
  \texttt{\{ehlavnova,ruder\}@google.com}}

\begin{document}
\maketitle

\begin{abstract}
A challenge towards developing NLP systems for the world's languages is understanding how they generalize to typological differences relevant for real-world applications. To this end, we propose M2C, a morphologically-aware framework for behavioral testing of NLP models. We use M2C to generate tests that probe models' behavior in light of specific linguistic features in 12 typologically diverse languages. We evaluate state-of-the-art language models on the generated tests. While models excel at most tests in English, we highlight generalization failures to specific typological characteristics such as temporal expressions in Swahili and compounding possessives in Finish. Our findings motivate the development of models that address these blind spots.\footnote{We make all code publicly available at \url{https://github.com/google-research/multi-morph-checklist}.}
\end{abstract}

\section{Introduction}

In natural language processing (NLP), there is a need to build systems that serve more of the world's approximately 6,900 languages. As one measure of linguistic diversity, the World Atlas of Language Structures \cite[WALS;][]{haspelmath2005world} records 192 linguistic features along which languages differ. These range from the order of subject, object, and verb \cite{wals-81} to the number of basic color categories \cite{wals-133}. Languages present in existing NLP datasets mostly lie in low-density regions of the space of possible typological features \cite{ponti-etal-2021-minimax}. In other words, many linguistic features that are common across the world's languages are not observed in languages that are the focus of NLP research.\footnote{For instance, while tone is present in around 80\% of African languages \cite{adebara-abdul-mageed-2022-towards}, few Indo-European languages can be considered tonal.}

\begin{figure}[t]

\begin{subfigure}[t]{\columnwidth}
    \centering
    \includegraphics[width=\columnwidth]{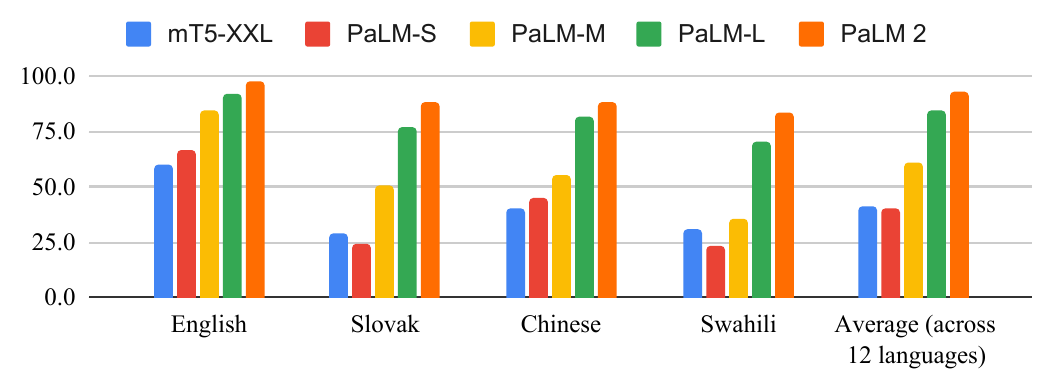}
\end{subfigure}
\begin{subfigure}[t]{\columnwidth}
    \centering
    \includegraphics[width=\columnwidth]{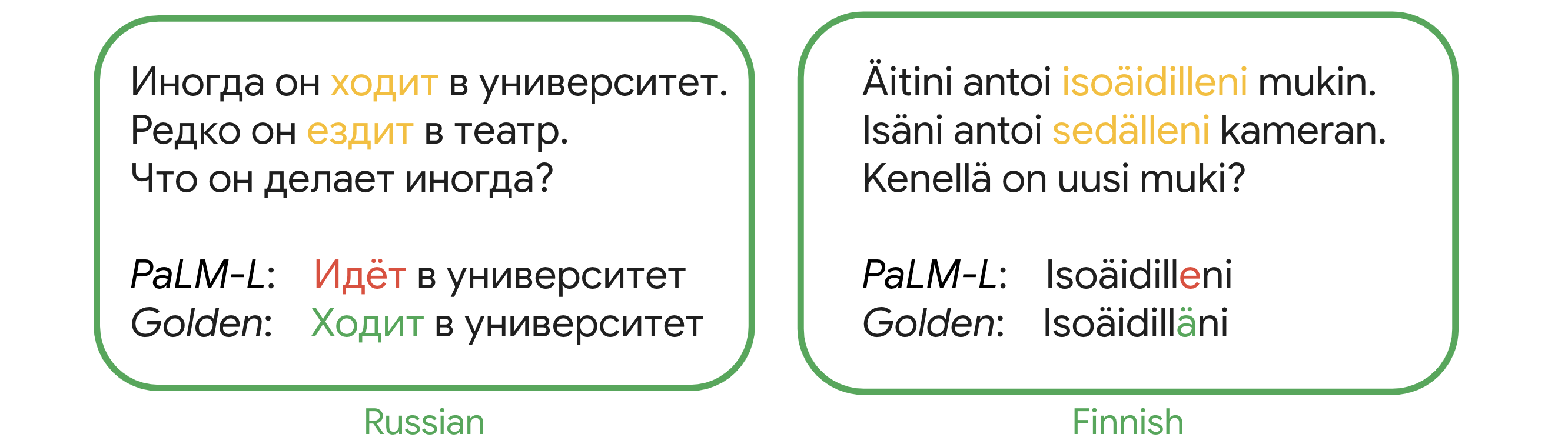}
\end{subfigure}
\caption{\emph{Top}: Comparison of state-of-the-art models on M2C tests in a selected set of languages. Models perform well on English but poorly on certain tests in other languages. \emph{Bottom}: Even the largest models fail on tests probing language-specific features, e.g., the distinction between habitual and one-time motion verbs in Russian (left) or possessives in Finnish (right); see Appendix \ref{app:errors_language-specific_features} for English glosses and additional examples.}
\label{fig:model_comparison}
\vspace{-6pt}
\end{figure}

It is thus important to investigate to which linguistic features models can generalize and where they face challenges. However, existing datasets do not allow for a fine-grained cross-lingual evaluation and mainly permit comparisons on a language level \cite{hu2020xtreme}. Prior studies 
focused on syntax and grammar through the lens of acceptability judgements \cite{ravfogel-etal-2018-lstm,ahmad-etal-2019-difficulties,mueller-etal-2020-cross,Papadimitriou2022accent}. While these enable the evaluation of what a model deems `natural' in a given language, it is often unclear how such biases relate to real-world applications of NLP technology.

\begin{figure*}
    \centering
    \includegraphics[width=\textwidth]{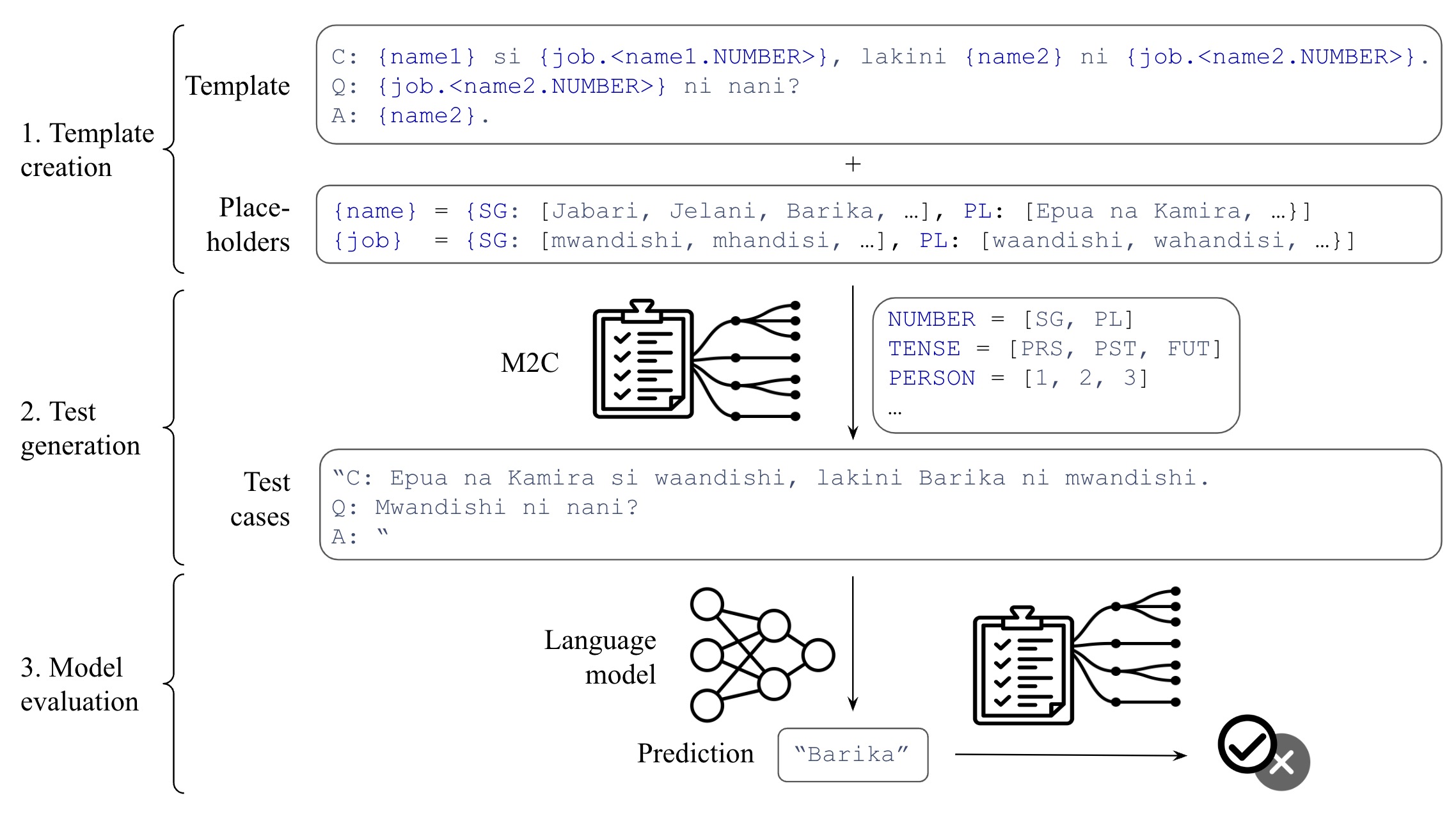}
    \caption{General workflow of using M2C for model evaluation. 1) Templates including context (C), question (Q), and answer (A) and placeholders for morphological features are created. 2) M2C is used to generate test cases. 3) A model is evaluated on the generated tests in a prompting setting and M2C is used to validate the predictions.}
    \label{fig:m2c_workflow}
\end{figure*}

We propose Multilingual Morphological Checklist (M2C) to enable the investigation of a broader set of cross-lingual differences in practical scenarios. Specifically, we create a morphologically-aware behavioral testing framework \cite{ribeiro-etal-2020-beyond} 
that allows for the specification of tests in a diverse set of languages. Using this framework, we design tests that probe model's behavior in light of specific capabilities and typological features in 12 typologically diverse languages. We focus on a question answering setting as it represents one of the most general and widely useful NLP applications \cite{Mccann2018} and enables zero-shot evaluation of models. We create tests that cover a diverse set of reasoning capabilities involving general linguistic features that are expressed differently across languages---negation, numerals, spatial and temporal expressions, and comparatives---as well as features unique to certain languages such as time in Swahili, measure words in Chinese, compounding possessives in Finnish, and motion verbs in Russian. We evaluate state-of-the-art language models on the generated tests in zero-shot and one-shot settings. Our findings shed light on generalization failures to specific typological features. For instance, all models struggle with time expressions in Swahili and measure words in Chinese. We show the workflow of using M2C, from template creation to model evaluation, in Figure \ref{fig:m2c_workflow}.

Our contributions are: 
(1) We create a new morphologically-aware multilingual behavioral testing framework. 
(2) We highlight linguistic features that are challenging in different languages.
(3) We design tests that probe model capabilities in light of practically relevant typological differences.
(4) We evaluate state-of-the-art language models on the generated tests.
(5) We shed light on the challenges posed by typological differences in multilingual scenarios.

\section{Related Work}

\paragraph{Perplexity} Perplexity is a standard measure of evaluating language model performance, which has also been used in multilingual settings \cite{gerz-etal-2018-language}. Besides being difficult to compare across segmentations, perplexity does not provide more fine-grained insights regarding model behavior \cite{meister-cotterell-2021-language}. Acceptability evaluations compare perplexity between minimal pairs of grammatical and ungrammatical sentences
\cite{linzen-etal-2016-assessing,warstadt-etal-2020-blimp-benchmark}. Such evaluations have been extended to other languages \cite{ravfogel-etal-2018-lstm,ahmad-etal-2019-difficulties,mueller-etal-2020-cross,xiang-etal-2021-climp,Papadimitriou2022accent}, which requires writing extensive language-specific grammars while the relevance of syntax biases in real-world applications remains unclear.

\paragraph{Evaluation of large models} Most benchmarks designed for evaluating large models focus on assessing their performance on a collection of complex tasks \cite{Wang2019superglue,hu2020xtreme,Hendrycks2021,gehrmann-etal-2021-gem,srivastava2022beyond}. However, such benchmarks are unable to highlight more fine-grained model limitations \cite{ethayarajh-jurafsky-2020-utility} and are outpaced by the development of new models. 

\paragraph{Behavioral testing} Behavioral testing sheds light on model capabilities via the design of simple targeted tasks. Early work such as bAbI \cite{weston2016babi} focused on toy tasks requiring simple reasoning capabilities while oLMpics \cite{talmor-etal-2020-olmpics} consisted of 8 short classification tasks for masked language models. Recently, LMentry \cite{Efrat2022lmentry} provides simple tests assessing fundamental generation capabilities. A common test bed is natural language inference \cite{naik-etal-2018-stress,mccoy-etal-2019-right} where analyses of reasoning types have been extended to other languages \cite{k-etal-2021-analyzing,joshi-etal-2020-taxinli,hartmann-etal-2021-multilingual} but require existing data. 

The CheckList framework \cite{ribeiro-etal-2020-beyond} enables the generation of behavioral tests for NLP models but its templates are English-centric. English Checklist tests have been extended to other languages via translation \cite{ruder-etal-2021-xtreme,k-etal-2022-multilingual}. Such approaches, however, struggle with comprehensively covering linguistic features specific to a language and are not able to easily represent morphological variation. Relatedly, \citet{jiang-etal-2020-x} create templates that integrate morphology for simple knowledge retrieval queries while \citet{kassner-etal-2021-multilingual} automatically translate knowledge retrieval queries into other languages. Compared to their approach, our framework allows for integrating morphology into a broader range of tests and is more scalable and flexible.

\section{CheckList}

CheckList \cite{ribeiro-etal-2020-beyond} relies on templates to generate a large amount of samples in order to evaluate models' behavior regarding different tasks and capabilities in a controlled manner. A template consists of a string with placeholders such as \ttag{first\_name} delimited by curly brackets, \eg \template{\ttag{first\_name} is \ttag{adj}}. The user provides a set of values for each placeholder, for instance, \ttag{first\_name} = \{\fillin{Michael}, \fillin{John}, ... \} and \ttag{adj} = \{\fillin{busy}, \fillin{friendly}, ... \}, which are used to populate the templates with their Cartesian product. The generated samples can then be applied to systematically test a model's performance in a specific setting.

\paragraph{Multilingual tests} CheckList has been designed for English and provides mainly English-specific functionality. For example, it matches indefinite articles with nouns based on their starting letter, \ie the placeholder \ttag{a:job} generates \exinline{a lawyer} and \exinline{an engineer}. As a consequence, CheckList is not capable of effectively generating tests in languages with richer morphology, which require maintaining agreement between multiple parts of the template---a feature that is beyond the scope of CheckList.

While multilingual tests can be generated by translating English tests \cite{ruder-etal-2021-xtreme,k-etal-2022-multilingual}, optionally including template extraction and human verification, such generated templates struggle with handling rich morphology. In addition, in order to systematically probe linguistic features specific to a language, it is crucial to be able to efficiently generate in-language tests from scratch.



\section{M2C Framework}

We propose the M2C (Multilingual Morphological Checklist) framework in order to enable the generation of tests in a broad set of languages, including languages with rich morphology.
A user provides a template as a string, a list of values for each placeholder, and an optional configuration dictionary in case of duplicate placeholders. The placeholder values can either be passed without inflections (for example, names in English) as a list of strings, or as a list of dictionaries with their corresponding inflected values. Each key of the dictionary is a feature combination (\eg \tfeat{MASC.PL}) and the value is the corresponding string (\textit{e.g.} \exinline{apples}). As such, each entity can have multiple inflections, for instance, in English \exinline{apple} and \exinline{apples}. We show the general M2C workflow in Figure \ref{fig:m2c_workflow}.

\paragraph{Morphological categories} Our library follows the UniMorph Schema representation \cite{sylak-glassman2016}, which decomposes morphology into 23 dimensions and over 212 features. For example, Gender is one dimension, which contains features such as Feminine (\textsc{fem}), Masculine (\textsc{masc}), and Neuter (\textsc{neut}).


The ability to indicate these dimensions using a clear codification allows us to describe both the value attributes given to placeholders and their dependence on one another. As an example, in order to differentiate between \exinline{Juliette est grande} and \exinline{Julien est grand} in French, it is necessary to ensure gender agreement between noun and adjective by including the Gender attribute in the template. To cover such functionality, we introduce a syntax describing the morphological dependence between placeholders: \ttag{X.<Y.D>} signifies that \tfeat{X} should have the same feature for dimension \tfeat{D} as \tfeat{Y}. In the above example, this is realized by \template{\ttag{first\_name} est \ttag{adj.<first\_name.GENDER>}}.

\paragraph{Language-specific dimensions} While initially relying on the UniMorph schema, we found cases where the existing dimensions are not sufficient to describe morphology of placeholders within the templates, which is especially necessary for dealing with exceptions. For instance, the trifold article distinction in Italian masculine gender---\textit{il treno}, \textit{l'hotel}, \textit{lo studente}---depends on whether the noun starts with a consonant, vowel or \textit{h}, or a specific consonant combination\footnote{\textit{gn, pn, ps, x, y, z, s} followed by another consonant or \textit{i} followed by a vowel.} respectively. In order to lexically encode such exceptions, we provide the ability to add dimensions, in this case \textsc{startswith}, which includes features \textsc{vow}, \textsc{cons}, and \textsc{cons2}. While the goal of M2C is not to be exhaustive, it should enable encoding a sufficient number of dimensions to allow the user to write templates for diverse use cases.\footnote{UniMorph defines a generic dimension `Language Specific features' with attributes \textsc{lgspec1}, .., \textsc{lgspecn}, which does not provide the clarity and flexibility of our setup.}

\paragraph{Advanced templating system}
To cover the variety of morphological phenomena, we designed a templating system with a rich syntax. When describing dependence rules, features can be added sequentially and are commutative, \eg \tfeat{<first\_name.GENDER.NUMBER>} is equivalent to \tfeat{<first\_name.NUMBER.GENDER>} where \tfeat{NUMBER} = \{\fillin{singular}, \fillin{plural}\}. Often, only two or three output values are necessary, which directly depend on a placeholder’s feature. We allow a simple expression to be passed directly in the template to make this rule explicit:
\vspace{-0.25in}
\begin{figure}[H]
\centering
\resizebox{\columnwidth}{!}{%
\begin{tabular}{ll}
\ttag{val\_1:placeholder.feature\_1 | ... | val\_n:placeholder.feature\_n},
\end{tabular}
}
\vspace{-0.3in}
\end{figure}
\noindent \eg \ttag{is:first\_name.SG|are:first\_name.PL}, which produces \exinline{is} for a singular \ttag{first\_name} and \exinline{are} for a plural one. 
Finally, we allow multiple placeholders with the same type, \eg \ttag{first\_name1} and \ttag{first\_name2}, to be populated by values of a common type, \ie \tfeat{first\_name}. In the case of multiple placeholders, 
we can provide a configuration for each placeholder type that specifies boolean repetition and order fields to, for instance, avoid having examples like \exinline{John and John} (repetition) or \exinline{John and Mary} and \exinline{Mary and John} (order). 

Manual enumeration of features and their corresponding values is a barrier to scaling. To circumvent this, we integrate UnimorphInflect \cite{anastasopoulos19emnlp}, which uses models trained on Unimorph data using the Unimorph Schema to generate inflections in 55 languages. As Unimorph models are imperfect---test accuracies range from 90\%+ in many languages to 23\% in Arabic---we envision a workflow where inflections are generated at scale using UnimorphInflect and then manually inspected by annotators for correctness. We expect the increase in productivity, and thus reduction in cost, to be significant by leveraging semi-automated as opposed to manual generation for languages with good performance.\footnote{In order to ensure high-quality tests for the experiments in \textsection \ref{sec:experiments}, we manually enumerate all relevant inflections.}

\paragraph{Answer validation} Most prior benchmarks for behavioral testing of language models have focused on classification tasks \cite{talmor-etal-2020-olmpics,ribeiro-etal-2020-beyond}. As M2C aims to support the evaluation of generative models using arbitrary templates, we implement functionality to match a range of outputs for each template, based on morphology, string matching and regex.\footnote{For each of the templates in \textsection \ref{sec:experiments}, we curate possible outputs and implement regex and functions capturing them.}


\paragraph{Summary} Overall, the M2C framework enables the systematic and controlled generation of high-quality tests at scale in a broad set of languages. As such, it occupies a middle ground between libraries such as SimpleNLG \cite{simple-nlg-2009} that generate high-quality data but require encoding each language-specific rule, and template expansion via generative language models \cite{Honovich2022}, which are highly scalable but less reliable and underperform on languages with limited data \cite{hu2020xtreme}. M2C enables modular design by allowing the addition of user-specified dimensions and features for specific templates and languages without requiring to encode all possible rules of a language. Furthermore, an advanced templating syntax and the semi-automatic generation of inflections may improve user productivity.


\begin{table*}[]
\centering
\resizebox{\textwidth}{!}{%
\begin{tabular}{c l l}

\toprule
Test & Template & Generated test \\ \midrule
Negation & 
\cellbreaks{

    .\placeholder{\ttag{job2.NOM.<name2.NUMBER.GENDER>}} \placeholder{\ttag{name2}}
    \RL{و} \placeholder{\ttag{job1.NOM.<name1.NUMBER.GENDER>}} \placeholder{\ttag{name1}} :C \\
    \RL{؟}\placeholder{\ttag{job1.ACC.<name2.NUMBER>.MASC}} \placeholder{\ttag{\RL{ليس}:name2.SG|\RL{ليسا}:name2.DU}} \RL{من } :Q \\
    .\placeholder{\ttag{name2}} :A
}
& \cellbreaks{
    \RL{أحمد مهندس وعمر كاتب.} :C \\
    \RL{من ليس مهندس؟} :Q \\
    \RL{عمر.} :A
}

\\ \verythinrule
Numerals & 
\cellbreaks{
    C: \foreignlanguage{russian}{На столе} \placeholder{\ttag{number1.<fruit1.GENDER>}} \placeholder{\ttag{fruit1.NOM.<number1.NUMBER>}}\\
    \hspace{1.2em}\foreignlanguage{russian}{и} \placeholder{\ttag{number2.<fruit2.GENDER>}} \placeholder{\ttag{fruit2.NOM.<number2.NUMBER>}}. \\
    \hspace{1.2em}\placeholder{\ttag{name}} \placeholder{\ttag{\foreignlanguage{russian}{съел}:name.MASC|\foreignlanguage{russian}{съела}:name.FEM}} \\
    \hspace{1.2em}\placeholder{\ttag{number3.<fruit1.GENDER>}} \placeholder{\ttag{fruit1.<ACC:number3.SG|NOM>.<number3.NUMBER>}}. \\
    Q: \foreignlanguage{russian}{Сколько} \placeholder{\ttag{fruit1.NOM.GTPL}} \foreignlanguage{russian}{на столе?} \\
    A: \placeholder{\ttag{\$diff(number1,number3)}}.
}
& \cellbreaks{
    C: \foreignlanguage{russian}{На столе три ягоды клубники и пять ананасов. } \\
    \hspace{1.2em}\foreignlanguage{russian}{Анна съела две ягоды клубники.} \\
    Q: \foreignlanguage{russian}{Сколько ягод клубники на столе?}\\
    A: \foreignlanguage{russian}{Одна.}
}
\\ \verythinrule

Spatial & 
\cellbreaks{
    C: \placeholder{\ttag{ART1.DEF.<obj1.NUMBER.STARTSWITH.GENDER>.TO\_CAPITALIZE}} \placeholder{\ttag{obj1}} e \\
    \hspace{1.2em}\placeholder{\ttag{ART2.DEF.<obj2.NUMBER.STARTSWITH.GENDER>}} \placeholder{\ttag{obj2}} sono \\
    \hspace{1.2em}\placeholder{\ttag{prep.<place.STARTSWITH.GENDER>}} \placeholder{\ttag{place}}. \\
    \hspace{1.2em}\placeholder{\ttag{name}} mette \placeholder{\ttag{ART2.DEF.<obj2.NUMBER.STARTSWITH.GENDER>}} \placeholder{\ttag{obj2}} sul pavimento. \\
    Q: \placeholder{\ttag{Dov'è:obj1.SG|Dove sono:obj1.PL}} \placeholder{\ttag{ART3.DEF.<obj1.NUMBER.STARTSWITH.GENDER>}} \placeholder{\ttag{obj1}}? \\
    A: \placeholder{\ttag{prep.<place.STARTSWITH.GENDER>.TO\_CAPITALIZE}} \placeholder{\ttag{place}}.
}
& \cellbreaks{
    C: Il libro e le penne sono accanto al tavolo. \\
    \hspace{1.2em}Leonardo mette le penne sul pavimento. \\
    Q: Dov'è il libro? \\
    A: Accanto al tavolo.
}
\\ \verythinrule

Temporal & 
\cellbreaks{
    C: \placeholder{\ttag{name1}} na \placeholder{\ttag{name2}} ni \placeholder{\ttag{job1.PL}} lakini \placeholder{\ttag{name1}} \\
    \hspace{1.2em}atabadilisha kazi na atakuwa \placeholder{\ttag{job2.SG}}. \\
    Q: \placeholder{\ttag{name1.TO\_CAPITALIZE}} atakuwa nani? \\
    A: \placeholder{\ttag{job2.SG.TO\_CAPITALIZE}}.
}
& \cellbreaks{
    C: Jabari na Jelani ni waandishi lakini \\
    \hspace{1.2em}Jabari atabadilisha kazi na atakuwa mwalimu \\
    Q: Jabari atakuwa nani? \\
    A: Mwalimu.
}
\\ \verythinrule

Comparative &
\cellbreaks{

    C:\begin{CJK*}{UTF8}{gbsn}如果\placeholder{\ttag{obj1}}\placeholder{\ttag{comp1.GT}}一点，\placeholder{\ttag{name}}会\placeholder{\ttag{act}}它。\end{CJK*} \\
    \hspace{1.2em}\begin{CJK*}{UTF8}{gbsn}如果\placeholder{\ttag{obj2}}\placeholder{\ttag{comp2.GT}}一点，\placeholder{\ttag{name}}会\placeholder{\ttag{act}}它。\end{CJK*} \\
    Q:\begin{CJK*}{UTF8}{gbsn}如果它不那么\placeholder{\ttag{comp1.LT}}，\placeholder{\ttag{name}}会\placeholder{\ttag{act}}什么？\end{CJK*} \\
    A: \:\:\placeholder{\ttag{obj1}}\begin{CJK*}{UTF8}{gbsn}。\end{CJK*}
}
& \cellbreaks{
    C:\begin{CJK*}{UTF8}{gbsn}如果公寓小一点，佳丽会买它。\end{CJK*}\\
    \hspace{1.2em}\begin{CJK*}{UTF8}{gbsn}如果电脑便宜一点，佳丽会买它。 \end{CJK*} \\
    Q:\begin{CJK*}{UTF8}{gbsn}如果它不那么大，佳丽会买什么？ \end{CJK*} \\
    A:\begin{CJK*}{UTF8}{gbsn}公寓。\end{CJK*}

}

\\ \bottomrule

\end{tabular}%
}
\caption{Templates including context (C), question (Q), and answer (A) with generated test examples for linguistic features in 
Arabic, Russian, Italian, Swahili, and Mandarin Chinese. Placeholders are defined within curly brackets with their morphological dependence.}
\label{tab:linguistic_feature_templates}
\end{table*}

\section{Capabilities and Typological Features}

\paragraph{Languages} We generate tests targeting capabilities and typological features in 12 typologically diverse languages: English (\en), Spanish (\es), Italian (\It), French (\fr), German (\de), Swedish (\sv), Finnish (\Fi), Slovak (\sk), Russian (\ru), Swahili (\sw), Mandarin Chinese (\zh), and Arabic (\ar).

Recent models have excelled at a wide range of tasks in English requiring a diverse set of reasoning and understanding capabilities \cite{Wang2019superglue, Hendrycks2021}. As most languages are morphologically richer than English, they encode the linguistic features representing such capabilities in more complex ways. The features we investigate are relevant in a variety of real-world applications including sentiment analysis \cite{wiegand-etal-2010-survey}, question answering \cite{dua-etal-2019-drop}, grounding \cite{kordjamshidi-etal-2020-representation}, reasoning with temporal change \cite{Lazaridou2021} and quantitative attributes \cite{elazar-etal-2019-large}.

We investigate capabilities and linguistic features present in all our investigated languages as well as linguistic features unique to certain languages. For each feature, we highlight differences in its cross-lingual instantiation and challenges for natural language understanding and generation. We create templates using the M2C framework to test a model's understanding of each capability and feature. We show a subset in Table \ref{tab:linguistic_feature_templates}. 

\subsection{Language-agnostic features}

\paragraph{Negation} 
In Indo-European languages, negation is often expressed via a separate particle such as \textit{not} (English), \textit{inte} (Swedish), etc.
In contrast, in Swahili, for instance, negation morphemes are fused with the verb root and thus harder to identify. For other negation terms such as \textit{kein} (German) models need to produce the correct agreement when generating text. In addition to gender and number agreement with the subject, Arabic negation takes up to five forms in singular, three forms in dual, and five forms in plural, \eg \<ليس>~(\texttt{SG.MASC}) and \<ليست>~(\texttt{SG.FEM}).

\paragraph{Numerals}

Models must be able to recognize and reason with numbers in their spelled-out and numerical forms across different writing and numeral systems, \eg \textit{seventeen} (English) and 17 (Western Arabic numerals) and \<سبعة عشر> and \<٧١> (Eastern Arabic numerals). 
For generation in Russian and Slovak, models must inflect the noun depending on the quantity of the object. Slovak, for instance, has separate inflections for quantities of one, two/three/four, and five and more, which also vary based on the object's animacy.

\paragraph{Spatial expressions}
In Russian, prepositions are associated with different cases, for example the instrumental case for \foreignlanguage{russian}{за} (\textit{behind})
and the prepositional case for \textit{on}. Such case agreement needs to be taken into account when generating text in Russian. Finnish, in addition to prepositions, follows a system of postpositions, which relate the location of one thing to another and require objects to be inflected in either partitive or genitive case. 

\paragraph{Temporal expressions}
%
Some languages with rich morphology such as Finnish and Swahili encode temporal expressions in less complex ways than their inflection-sparser counterparts. In Swahili, verbal structure follows a simple compounding schema of subject marker + tense marker + verb, e.g. \textit{a-na-soma} (\textit{he reads}) or \textit{u-ta-soma} (\textit{you will read}). 

\paragraph{Comparatives}
Commonly, comparatives are expressed by a suffix or using a quantifier, \eg \textit{more/less}. Spanish and French follow the latter approach by placing \textit{más/menos} and \textit{plus/moins} before the adjective with only a few standard exceptions. 
On the other hand, in Finnish, for example, the formation of comparatives follows a complex system of rules for compounding that includes categories depending on the endings of adjectives and a suffix \textit{mpi}. 

\subsection{Language-specific features}


\paragraph{Time in Swahili} In many languages, the day is divided into two periods: a.m. and p.m., with the daily cycle starting at midnight (0:00) and running through noon (12:00). In Swahili, time is based on sunset and sunrise, defined to be 6 pm and 6 am respectively in standard time. For example, 11.30 am in standard time is 5.30 in the morning in Swahili time. Understanding different time systems is key not only for in-language reasoning but also for cross-lingual applications.

\begin{table}[]
\centering
\begin{tabular}{ll}
\toprule
& Prompt: Svara på frågan. \\ \midrule
\parbox[t]{2mm}{\rotatebox[origin=c]{90}{\begin{tabular}[c]{@{}c@{}}Spatial\end{tabular}}} & \begin{tabular}[c]{@{}l@{}}
Kontext: Pennan är under stolen \\ och telefonen är på fönstret.\\ Fråga: Var är telefonen?\\ Svar: På fönstret\\ \textbf{Kontext: Boken är under soffan} \\ \textbf{och pennan är på hyllan.}\\ \textbf{Fråga: Var är pennan?}\\ \textbf{Svar:}\end{tabular} \\ \midrule
\end{tabular}%
\caption{Zero-shot and few-shot prompt example in Swedish spatial template. The zero-shot prompt only includes the information in bold while the one-shot prompt also includes the additional exemplar.}
\label{tab:zero_shot_one_shot_examples}
\end{table}


\paragraph{Possessives in Finnish}

Compounding in Finnish along with its system of 15 cases is one of the most challenging aspects of the language. One relevant feature are the possessive suffixes, which attach to the stem of nouns, \eg \textit{koulu} (\textit{school}) becomes \textit{kouluni} (\textit{my school}) and \textit{koulumme} (\textit{our school}).
Possession is expressed via a suffix \textit{-lla}, which compounds with other suffixes, \eg \textit{siskollani} (\textit{my sister has}), which must be correctly inflected by models in order to achieve the intended meaning.

\paragraph{Measure words in Mandarin Chinese}
Another language specific-feature are measure words in Mandarin Chinese, which include over 150 cases and are used for different types of objects depending on their characteristics, \begin{CJK*}{UTF8}{gbsn} \eg ``本'' for books, ``双'' for pairs, or ``辆'' for vehicles.\end{CJK*} 

\paragraph{Motion verbs in Russian}
In most Slavic languages, motion verbs are a challenging concept as they behave differently than other verb categories. While most verbs have two forms (imperfective and perfective), motion verbs have three forms: one perfective form and two imperfective forms. Of the imperfective forms, the definite form indicates unidirectional or current one-time motion while the indefinite form represents multi-directional or habitual motion.





\begin{table*}[t!]
\centering
\resizebox{\textwidth}{!}{%
\begin{tabular}{lccccccccccccc}
\toprule
 & \en & \es & \It & \fr & \de & \sv & \Fi & \sk & \ru & \zh & \sw & \ar & Avg. \\ \midrule
mT5-XXL & 59.6 & 32.0 & 43.9 & 41.4 & 50.4 & 39.3 & 44.8 & 28.5 & 39.1 & 40.0 & 30.6 & 52.1 & 41.8 \\
PaLM-S & 66.5 & 38.9 & 36.6 & 47.9 & 47.1 & 53.3 & 39.8 & 23.9 & 33.9 & 44.7 & 23.4 & 29.4 & 40.4 \\
PaLM-M & 84.5 & 70.9 & 60.1 & 78.2 & 71.8 & 66.2 & 53.5 & 50.6 & 54.0 & 55.1 & 35.1 & 48.8 & 60.7 \\
PaLM-L & 92.5 & 89.5 & 89.2 & 92.0 & 86.7 & 90.7 & 87.4 & 76.8 & 80.5 & 82.0 & 70.6 & 78.1 & 84.7 \\
PaLM 2 & \textbf{98.1} & \textbf{98.2} & \textbf{93.6} & \textbf{98.3} & \textbf{95.0} & \textbf{97.0} & \textbf{88.7} & \textbf{88.5} & \textbf{93.1} & \textbf{88.3} & \textbf{83.9} & \textbf{91.2} & \textbf{92.8} \\
\bottomrule
\end{tabular}%
}
\caption{Average accuracy (in \%) of different models on the generated tests in a zero-shot setting.}
\label{tab:main-language-model-comparison}
\end{table*}


\section{Experiments} \label{sec:experiments}

\begin{table*}[]
\centering
\resizebox{\textwidth}{!}{%
\begin{tabular}{c c c c c c c c c c c c c c c c c}
\toprule
& Test type & Model & \en & \es & \It & \fr & \de & \sv & \Fi & \sk & \ru & \zh & \sw & \ar & Avg. \\ \midrule

\parbox[t]{2mm}{\multirow{4}{*}{\rotatebox[origin=c]{90}{\begin{tabular}[c]{@{}c@{}}Negation\end{tabular}}}} & \multirow{2}{*}{In context} & 
mT5-XXL & 80.7 & 72.8 & 85.5 & 80.2 & 63.1 & 55.8 & 84.4 & 31.8 & 45.3 & 30 & 33.7 & 43.1 & 56.9 \\
& & PaLM 2 & \textbf{99.9} & \textbf{100} & \textbf{98.4} & \textbf{100} & \textbf{100} & \textbf{100} & \textbf{100} & \textbf{100} & \textbf{100} & \textbf{90.1} & \textbf{100} & \textbf{92.3} & \textbf{98.3} \\
& \multirow{2}{*}{In question} & 
mT5-XXL & 19.1 & 30.1 & 23.4 & 25.1 & 36.1 & 20.6 & 19.7 & 16.7 & 9.6 & 5.2 & 3.7 & 58.2 & 22.6 \\
& & PaLM 2 & \textbf{100} & \textbf{100} & \textbf{98.9} & \textbf{100} & \textbf{99.8} & \textbf{99.3} & \textbf{100} & \textbf{100} & \textbf{100} & \textbf{76.6} & \textbf{99.6} & \textbf{95.1} & \textbf{97.2} \\ \verythinrule

\parbox[t]{2mm}{\multirow{4}{*}{\rotatebox[origin=c]{90}{\begin{tabular}[c]{@{}c@{}}Numerals\end{tabular}}}} & 
\multirow{2}{*}{Addition} & 
mT5-XXL & 0.4 & 0.2 & 2.3 & 2 & 1.7 & 1.6 & 0 & 0 & 0 & 0 & 0.1 & 42.6 & 4.6 \\
& & PaLM 2 & \textbf{96.1} & \textbf{100} & \textbf{68.7} & \textbf{99.7} & \textbf{96.5} & \textbf{100} & \textbf{100} & \textbf{99.9} & \textbf{96.9} & \textbf{66.5} & \textbf{94.5} & \textbf{79.3} & \textbf{91.1} \\
& \multirow{2}{*}{Subtraction} & 
T5-XXL & 33.4 & 21.5 & 24.2 & 22.2 & 33 & 31.3 & 26.8 & 19.8 & 12.9 & 23 & 5.9 & 32.1 & 23.0 \\
& & PaLM 2 & \textbf{95} & \textbf{92.4} & \textbf{90} & \textbf{93.6} & \textbf{93.6} & \textbf{89.1} & \textbf{87.5} & \textbf{88.4} & \textbf{93.6} & \textbf{81.2} & \textbf{68.7} & \textbf{87.4} & \textbf{87.8} \\\verythinrule

\parbox[t]{2mm}{\multirow{4}{*}{\rotatebox[origin=c]{90}{\begin{tabular}[c]{@{}c@{}}Spatial\end{tabular}}}} & 
\multirow{2}{*}{Prepositions} & 
mT5-XXL & 98.8 & 28 & 51.4 & 40.2 & 78.3 & 59.6 & 27.6 & 51.3 & 49.5 & 99.9 & 52.8 & 74.4 & 55.7 \\
& & PaLM 2 & \textbf{100} & \textbf{100} & \textbf{94.8} & \textbf{100} & \textbf{100} & \textbf{100} & \textbf{100} & \textbf{100} & \textbf{99.9} & \textbf{100} & \textbf{100} & \textbf{98.7} & \textbf{99.4} \\
& \multirow{2}{*}{Position} & 
mT5-XXL & 90.9 & 15 & 74.5 & 61.1 & 95.2 & 35.1 & 60.3 & 29 & 50 & \textbf{100} & 49 & 65.3 & 57.7 \\
& & PaLM 2 & \textbf{100} & \textbf{100} & \textbf{99.9} & \textbf{100} & \textbf{100} & \textbf{100} & \textbf{99} & \textbf{100} & \textbf{99.9} & \textbf{100} & \textbf{46.7} & \textbf{91.0} & \textbf{94.2} \\ \verythinrule

\parbox[t]{2mm}{\multirow{4}{*}{\rotatebox[origin=c]{90}{\begin{tabular}[c]{@{}c@{}}Temporal\end{tabular}}}} & \multirow{2}{*}{Past} & 
mT5-XXL & 86.3 & 27.8 & 44.4 & 62.1 & 50.4 & 77.5 & 78.7 & 61.7 & 93.1 & 81.1 & 35.2 & 68.9 & 61.9 \\
& & PaLM 2 & \textbf{99.3} & \textbf{100} & \textbf{89.8} & \textbf{100} & \textbf{86.8} & \textbf{100} & \textbf{100} & \textbf{83.5} & \textbf{96.9} & \textbf{96.7} & \textbf{62.9} & \textbf{96.2} & \textbf{92.1} \\ 

& \multirow{2}{*}{Future} & 
mT5-XXL & 85.7 & 79.8 & 48.4 & 56.9 & 55.3 & 55 & 62.2 & 38.3 & 93.5 & 52.7 & 39 & 58.7 & 58.2 \\
& & PaLM 2 & \textbf{100} & \textbf{100} & \textbf{100} & \textbf{100} & \textbf{100} & \textbf{100} & \textbf{100} & \textbf{95} & \textbf{99.1} & \textbf{100} & \textbf{100} & \textbf{99.8} & \textbf{99.4} \\ \verythinrule

\parbox[t]{2mm}{\multirow{4}{*}{\rotatebox[origin=c]{90}{\begin{tabular}[c]{@{}c@{}}Comparative\end{tabular}}}} & 
\multirow{2}{*}{Standard} & 
mT5-XXL & 58.1 & 44 & 37.3 & 48.7 & 45.3 & 28.3 & 60 & 31.3 & 17.3 & 7.7 & 51.7 & 45.3 & 37.9 \\
& & PaLM 2 & \textbf{100} & \textbf{97.7} & \textbf{100} & \textbf{100} & \textbf{100} & \textbf{100} & \textbf{100} & \textbf{100} & \textbf{100} & \textbf{99.3} & \textbf{100} & \textbf{100.0} & \textbf{99.7} \\ 
& \multirow{2}{*}{Conditional} & 
mT5-XXL & 42.4 & 1.1 & 47.8 & 15.8 & 45.5 & 28.1 & \textbf{28.7} & 4.7 & 19.8 & 0 & 35.2 & 32.1 & 23.5 \\
& & PaLM 2 & \textbf{90.6} & \textbf{92.1} & \textbf{95.1} & \textbf{89.4} & \textbf{73.3} & \textbf{81.7} & 0 & \textbf{18.1} & \textbf{44.2} & \textbf{72.7} & \textbf{66.3} & \textbf{72.1} & \textbf{64.1}  \\ \bottomrule
\end{tabular}
}
\caption{Accuracy (in \%) of mT5-XXL and PaLM 2 on the generated tests in a zero-shot setting.}
\label{tab:mt5_palm2_results}
\end{table*}

\paragraph{Experimental setting} We evaluate models on the generated tests in a question answering setting as can be seen in Figure \ref{fig:m2c_workflow}. Each test consists of a context, a question, and an answer that needs to be predicted by the model. For each template, we generate 2,000 test examples on which the model is evaluated. A model's performance on a template is its accuracy of predicting a valid answer for a test averaged across all tests of the template. 

We evaluate models in both zero-shot and one-shot settings for each capability and language. In the one-shot setting, a test randomly generated using the same template is used as the exemplar. This simplifies the task in two ways: i) it provides the model with a clear format for generating the answer and may enable the model to infer the answer's relationship to the rest of the template. While we conduct one-shot experiments to show the impact of additional instructions, zero-shot evaluation is the only setting that fully tests the model's understanding and generative capabilities independent of confounders such as the exemplar choice \cite{zhao2021calibrate}, in line with prior work on behavioral testing \cite{ribeiro-etal-2020-beyond,Efrat2022lmentry}. We provide an example of both settings in Table \ref{tab:zero_shot_one_shot_examples}.


\paragraph{Models} We evaluate five state-of-the-art pre-trained language models of different sizes: an LM-adapted version \cite{vu-etal-2022-overcoming} of mT5-XXL \cite[13B parameters;][]{xue-etal-2021-mt5}; PaLM-S (8B parameters), PaLM-M (62B parameters), and PaLM-L \cite[540B parameters;][]{chowdhery2022palm}; and PaLM 2 \cite{palm2}. All models have been trained on large amounts of web text but have not been otherwise fine-tuned for instruction-following or few-shot learning.

\paragraph{Generation} Predictions are generated using greedy decoding with a temperature of 0 and a maximum of 20 decoding steps.

\section{Results}

\subsection{Performance across Languages}

We show the average results across tests covering language-agnostic features across languages and models in Table \ref{tab:main-language-model-comparison}. We present the detailed results across test types for mT5-XXL and PaLM 2 in Table \ref{tab:mt5_palm2_results} and for PaLM-S, PaLM-M, and PaLM-L in Appendix \ref{app:zero-shot}. We show results on language-specific features for all models in Table \ref{tab:language-specific-feature-tests}.

\paragraph{M2C tests are challenging, particularly for smaller models and for certain languages.} mT5-XXL and PaLM-S achieve comparatively poor performance on average across languages. While performance is highest for English, across the other languages both models only pass at most 50\% of tests---and less than a third for Slovak (\sk), Swahili (\sw), and Arabic (\ar) for PaLM-S. These results highlight that the tests generated with M2C are challenging for the majority of state-of-the-art models and demonstrate that a clear gap between performance on English and performance in other languages remains for most models.

\paragraph{Competence with language-agnostic features emerges at scale.} We observe a 20 point improvement in average performance from PaLM-S to PaLM-M to PaLM-L, highlighting that model robustness to linguistic features improves with scale. The strongest model, PaLM 2, reaches almost perfect performance on English and on the Indo-European languages. Compared to PaLM-L, PaLM 2 achieves the largest improvements on Slovak, Russian, Swahili, and Arabic. On Finnish, Slovak, Chinese, and Swahili average performance of PaLm 2 is still below 90\%, however, indicating that there is headroom left in terms of competence with regard to language-agnostic features for even the strongest current models.

\subsection{Performance across Linguistic Features}

\paragraph{Language-agnostic features} The most challenging test types for mT5-XXL and PaLM 2 in Table \ref{tab:mt5_palm2_results} are numerals and comparatives. mT5 performs poorly on addition and only slightly better on subtraction while PaLM 2 achieves around 90\% performance on most languages. On comparatives, both models have more difficulty in the conditional case. While PaLM 2 passes negation tests with almost perfect accuracy across different languages, mT5 displays reduced performance, particularly when the question is negated and for non-Indo-European languages. This highlights that robust reasoning with negation only emerges at scale. On spatial and temporal tests, mT5 achieves reasonable performance in most languages, while PaLM 2 achieves perfect performance in most cases and only underperforms in Swahili.

\paragraph{Language-specific features} We show the results on the language-specific feature tests in Table \ref{tab:language-specific-feature-tests}. All models have acquired a reasonable ability to distinguish between different forms of motion verbs in Russian. Small and medium-sized models generally fail to reason with compounding possessives in Finnish and time expressions in Swahili while all models are unable to perfectly employ the correct measure words in Chinese, despite it being a high-resource language. Similarly, even PaLM 2 is unable to correctly reason with time expressions in Swahili. We show examples of errors in model predictions for each test type together with English glosses in Appendix \ref{app:errors_language-specific_features}.




\begin{table}[]
\centering
\resizebox{\columnwidth}{!}{%
\begin{tabular}{lccccc}
\toprule
 & \Fi & \ru & \zh & \sw & Avg. \\ \midrule
mT5-XXL & 1.2 & 62.6 & 38.8 & 0 & 25.7 \\
PaLM-S & 3.6 & 68.1 & 5.1 & 0 & 19.2 \\
PaLM-M & 12.4 & 86.9 & 61.4 & 0 & 40.2 \\
PaLM-L & 63.4 & 90 & 71.6 & 13.6 & 59.7 \\
PaLM 2 & \textbf{98.7} & \textbf{99.4} & \textbf{77.5} & \textbf{69} & \textbf{86.2} \\
\bottomrule
\end{tabular}%
}
\caption{Accuracy (in \%) on tests testing language-specific features: time (Swahili), possessives (Finish), measure words (Chinese), motion verbs (Russian).}
\label{tab:language-specific-feature-tests}
\end{table}

\subsection{Evaluating Morphological Correctness}

The generated tests focus on evaluating a model's understanding capability with regard to specific capabilities and linguistic features. As the linguistic features are often expressed via morphology, we additionally calculate the fraction of errors due to morphology in the models' output for the tests with morphological variation in the answer. This enables us to assess a model's ability to generate morphologically correct forms. For instance, in Slovak, a model must generate the correct accents and suffixes, \eg it is an error if the model predicts the \textit{Trináste} (13\textit{th}) instead of \textit{Trinásť} (13). We automatically identify and manually curate these errors for PaLM-L and report the proportion of morphology-related errors for a subset of tests and languages in Table \ref{tab:morphological_errors}. We show examples of errors in model predictions that are due to morphology in Appendix \ref{app:morphological_errors}.

\begin{table}[]
\centering



\begin{tabular}{c c c c c c c c c c c c c c}
\toprule
& Languages & \Fi & \sk & \ru \\ \midrule

\parbox[t]{6mm}{\multirow{2}{*}{\rotatebox[origin=c]{90}{\begin{tabular}[c]{@{}c@{}}Neg-\\ation\end{tabular}}}} 
& \begin{tabular}[c]{@{}c@{}}In context\end{tabular}
& 31.6   & \colorbox{lightgray}{45.7}  & 27.6     \\
& \begin{tabular}[c]{@{}c@{}}In question\end{tabular}
& 10   & 51.8  & 3.2   \\ \midrule

\parbox[t]{6mm}{\multirow{2}{*}{\rotatebox[origin=c]{90}{\begin{tabular}[c]{@{}c@{}}Num-\\erals\end{tabular}}}} & 
\begin{tabular}[c]{@{}c@{}}Addition\end{tabular}
& 8   & 16.2  & 4.2\\
& \begin{tabular}[c]{@{}c@{}}Subtraction\end{tabular}
& 12.4   & 30  & \colorbox{lightgray}{11.8}  \\ \midrule

\parbox[t]{6mm}{\multirow{2}{*}{\rotatebox[origin=c]{90}{\begin{tabular}[c]{@{}c@{}}Spa-\\tial\end{tabular}}}} & 
\begin{tabular}[c]{@{}c@{}}Prepositions\end{tabular}
& \colorbox{lightgray}{7.8}  & 8.2  & 0 \\
& \begin{tabular}[c]{@{}c@{}}Position\end{tabular}
& 0   & 0  & 0.1  \\ \midrule

\parbox[t]{6mm}{\multirow{2}{*}{\rotatebox[origin=c]{90}{\begin{tabular}[c]{@{}c@{}}Temp-\\oral\end{tabular}}}} & 
\begin{tabular}[c]{@{}c@{}}Past\end{tabular}
& 0   & \colorbox{lightgray}{21.8}  & 39.8 \\
& \begin{tabular}[c]{@{}c@{}}Future\end{tabular}
& 0   & 8.3  & 0  \\ \midrule

\parbox[t]{6mm}{\multirow{2}{*}{\rotatebox[origin=c]{90}{\begin{tabular}[c]{@{}c@{}}Comp-\\arative\end{tabular}}}} & 
\begin{tabular}[c]{@{}c@{}}Standard\end{tabular}
& 0   & 0  & 0  \\
& \begin{tabular}[c]{@{}c@{}}Conditional\end{tabular}
& \colorbox{lightgray}{4.5}   & 3.2  & 25.6 \\
\bottomrule
\end{tabular}
\caption{Percentage of morphological errors (in \%) by PaLM-L on the generated tests with zero-shot setting. Example erroneous predictions corresponding to \colorbox{lightgray}{highlighted cells} are in Appendix \ref{app:morphological_errors}.}
\label{tab:morphological_errors}
\end{table}

For certain tests with morphological variation in the answer, a non-negligible fraction of errors are due to producing incorrect morphological forms. For negation in Slovak, around half of PaLM-L's errors are due to morphology such as an incorrect use of diacritics or suffixes, highlighting a weakness of subword-based models. For numerical reasoning, models frequently produce incorrectly inflected numerals. Similarly, models generate outputs with an incorrect case or number for tests related to spatial and temporal expressions and comparatives.


\subsection{One-shot Evaluation}

We show one-shot results for all models in Appendix \ref{app:one-shot}. The one-shot setting generally improves results as it allows the model to infer the format of the answer and potentially its relationship to the rest of the template. Improvements are larger for smaller models, which benefit more from information about the template. Nevertheless, even in this setting models are unable to achieve perfect accuracy across all languages. Reasoning with numerals and comparatives are still challenging for most models while improvements on numerals are also relatively smaller than on other test types. Models struggle particularly in Swahili across different test types. Overall, these results demonstrate that even in one-shot settings, large language models are not able to systematically generalize to certain typological features in multilingual settings.


\section{Conclusion}
In this paper, we have introduced M2C, a multilingual morphological framework for targeted behavioral evaluation of language-specific capabilities. As world languages present different challenges, M2C aims to provide flexibility in defining a suitable templating system with its individual dimensions and features. We have conducted experiments on state-of-the-art large language models, highlighted typological features that models struggle with, and quantified errors occurring due to morphology. We hope M2C inspires further research focused on tackling typological and morphological challenges with large language models. 

\section*{Acknowledgements}

We thank Jialu Liu, Jiaming Shen, and Jonas Pfeiffer for helpful feedback on a draft of this paper.

\section*{Broader Impact Statement}

\paragraph{Accessibility} Our new behavioral testing framework enables the generation of tests that incorporate morphology, which makes the systematic and fine-grained evaluation of NLP models more accessible across a diverse set of languages. For many such languages, it was previously not feasible to gain a fine-grained understanding of a model's capabilities.

\paragraph{Risks} Risks are limited and mainly relate to obtaining a biased view of a capability due to the use of limited templates.

\paragraph{Limitations} The creation of templates still requires native speaker expertise and an understanding of a language's grammar. Morphological inflection models are imperfect so morphological forms may need to be enumerated to ensure high-quality tests. We leave model-in-the-loop template creation and improving morphological inflection models for future work. While we design representative templates with thousands of permutations for each capability, a larger set of templates and arguments may be necessary to ensure a comprehensive coverage.


\bibliography{main}
\bibliographystyle{acl_natbib}

\appendix

\section{Zero-shot Results} \label{app:zero-shot}

We show zero-shot results for PaLM-S, PaLM-M, and PaLM-L across different tests and languages in Table \ref{tab:zero-shot-comparison-palm-models-app}.

\begin{table*}[]
\centering
\resizebox{\textwidth}{!}{%
\begin{tabular}{lllccccccccccccc}
\toprule
 & Test type & Model & \en & \es & \It & \fr & \de & \sv & \Fi & \sk & \ru & \zh & \sw & \ar & Avg. \\ \midrule
\parbox[t]{2mm}{\multirow{6}{*}{\rotatebox[origin=c]{90}{Negation}}} & \multirow{3}{*}{In context} & PaLM-S & 47.6 & 31.6 & 33.7 & 40.6 & 42.2 & 31.1 & 29.8 & 29.2 & 20.1 & 47.7 & 37.6 & 30.3 & 35.1 \\
 &  & PaLM-M & 97.6 & 61.3 & 75.1 & 89.1 & 71 & 89.2 & 71.4 & 83.5 & 44.6 & 50.4 & 45.4 & 40.1 & 68.2 \\
 & & PaLM-L & 99.9 & 99 & 99.3 & 99.2 & 99.7 & 100 & 99.7 & 95.2 & 90.5 & 90.2 & 97.6 & 91.2 & 96.8 \\
 & \multirow{3}{*}{In question} & PaLM-S & 56.5 & 30.4 & 44.7 & 29.2 & 37.2 & 38.8 & 39.9 & 30.9 & 46 & 49.8 & 9.4 & 26.9 & 36.6 \\
 &  & PaLM-M & 85.6 & 43.1 & 58.2 & 61.8 & 57.9 & 60.9 & 36.3 & 51.9 & 45.3 & 34.4 & 62.8 & 36.4 & 52.9 \\
 & & PaLM-L & 99.7 & 90.1 & 94.2 & 95.2 & 99.6 & 99.3 & 99.3 & 90 & 92.6 & 65 & 73.3 & 74.2 & 89.4 \\ \verythinrule
\parbox[t]{2mm}{\multirow{6}{*}{\rotatebox[origin=c]{90}{Numerals}}} & \multirow{3}{*}{Addition} & PaLM-S & 66.7 & 43.5 & 33.5 & 36.9 & 43.3 & 48.3 & 3.5 & 1.9 & 1.5 & 1 & 0 & 4.8 & 23.7 \\
 &  & PaLM-M & 77.2 & 55.8 & 17.2 & 64.3 & 56.8 & 10.4 & 15.9 & 9.3 & 22.7 & 36.1 & 2.8 & 12.5 & 31.8 \\
 & & PaLM-L & 96.5 & 92.6 & 94.1 & 97.9 & 98.2 & 87 & 61.7 & 58.8 & 74.6 & 82.5 & 47.3 & 59.7 & 79.2 \\
 & \multirow{3}{*}{Subtraction} & PaLM-S & 47.8 & 8.3 & 18.1 & 27.1 & 32 & 4 & 17.4 & 4 & 4.8 & 20 & 0 & 11.4 & 16.2 \\
 &  & PaLM-M & 44.5 & 38.2 & 36.4 & 65.4 & 46.2 & 35.4 & 27.5 & 17.6 & 34.1 & 59 & 0.1 & 31.9 & 36.4 \\
 & & PaLM-L & 93.6 & 92.4 & 77.9 & 93 & 62.2 & 86 & 87.9 & 77.2 & 90.4 & 86.6 & 33 & 53.1 & 77.8 \\ \verythinrule
\parbox[t]{2mm}{\multirow{6}{*}{\rotatebox[origin=c]{90}{Spatial}}} & \multirow{3}{*}{Prepositions} & PaLM-S & 84.9 & 73.6 & 52 & 75.9 & 86.2 & 88.6 & 41.5 & 31.5 & 30.1 & 87.4 & 14.1 & 31.9 & 58.1 \\
 &  & PaLM-M & 99 & 99 & 50 & 94.5 & 93.3 & 77 & 67.1 & 72.2 & 87.9 & 96.2 & 88.7 & 65.3 & 82.5 \\
 & & PaLM-L & 99.9 & 95.8 & 96.9 & 99.7 & 100 & 99.5 & 93.3 & 88.2 & 99 & 100 & 99.4 & 96 & 97.3 \\
 & \multirow{3}{*}{Position} & PaLM-S & 54.6 & 49.4 & 38.6 & 44.8 & 59 & 58.5 & 44.9 & 14.9 & 11.8 & 49.2 & 20.9 & 29.1 & 39.6 \\
 &  & PaLM-M & 65.9 & 73.3 & 42 & 56 & 62.5 & 57 & 46.7 & 31.4 & 36.7 & 62.4 & 18.9 & 42.8 & 49.6 \\
 & & PaLM-L & 61.1 & 58.7 & 74.6 & 55.1 & 73.7 & 74.6 & 88.1 & 66.9 & 41 & 89.2 & 23.7 & 74.7 & 65.1 \\ \verythinrule
\parbox[t]{2mm}{\multirow{6}{*}{\rotatebox[origin=c]{90}{Temporal}}} & \multirow{3}{*}{Past} & PaLM-S & 81.9 & 14.1 & 21.5 & 36.5 & 31.1 & 95 & 74.5 & 26 & 90.5 & 80.5 & 38.4 & 41.2 & 52.6 \\
 &  & PaLM-M & 99.6 & 94.5 & 92.9 & 95.8 & 94.4 & 92.1 & 98 & 93.8 & 94.7 & 96.3 & 40 & 67.8 & 88.3 \\
 & & PaLM-L & 100 & 98.5 & 84.4 & 99.8 & 99.9 & 95.3 & 100 & 95.9 & 95.6 & 99.8 & 93.6 & 94.3 & 96.4 \\
 & \multirow{3}{*}{Future} & PaLM-S & 92 & 36.3 & 15 & 77.5 & 32.6 & 95.5 & 60.2 & 20.6 & 89.1 & 86.8 & 58.1 & 51.2 & 59.6 \\
 &  & PaLM-M & 99.9 & 94.2 & 93.6 & 94.8 & 91.5 & 95.4 & 98.6 & 59.9 & 72.2 & 88.4 & 30.4 & 69.3 & 82.4 \\
 & & PaLM-L & 100 & 98.4 & 98.9 & 96.2 & 100 & 99.2 & 99.8 & 81.2 & 91.4 & 95.6 & 98.4 & 92.4 & 96.0 \\ \verythinrule
\parbox[t]{2mm}{\multirow{6}{*}{\rotatebox[origin=c]{90}{Comparative}}} & \multirow{3}{*}{Standard} & PaLM-S & 75.3 & 57.3 & 69.3 & 96.7 & 73.7 & 71.7 & 79.7 & 62 & 43.3 & 25 & 55.7 & 38.1 & 62.3 \\
 &  & PaLM-M & 92.3 & 69 & 80.7 & 83 & 82.7 & 77 & 58 & 71.7 & 69.7 & 26.3 & 61.3 & 73.1 & 70.4 \\
 & & PaLM-L & 100 & 87.3 & 100 & 98.7 & 100 & 100 & 99.3 & 100 & 98.7 & 86 & 99.3 & 91.6 & 96.7 \\
 & \multirow{3}{*}{Conditional} & PaLM-S & 57.2 & 44.2 & 39.2 & 13.4 & 33.9 & 1.3 & 6.4 & 17.6 & 1.7 & 0 & 0.1 & 29.5 & 20.4 \\
 &  & PaLM-M & 82.9 & 80.6 & 55.2 & 77 & 62.1 & 67.7 & 15.1 & 14.2 & 32.3 & 1.6 & 0.6 & 49.2 & 44.9 \\
 & & PaLM-L & 73.9 & 82.4 & 71.8 & 85.3 & 33.6 & 65.8 & 44.8 & 14.3 & 31 & 25 & 40.4 & 53.6 & 51.8 \\
 \bottomrule
\end{tabular}%
}
\caption{Accuracy (in \%) of PaLM-S, PaLM-M, and PaLM-L on generated tests in a zero-shot setting.}
\label{tab:zero-shot-comparison-palm-models-app}
\end{table*}

\section{Examples of Errors on Language-specific Feature Tests} \label{app:errors_language-specific_features}

We show examples of errors on language-specific feature tests with PaLM-L together with English glosses in Table \ref{tab:language-specific_feature_prediction_errors}.

\begin{table*}[]
\centering
\resizebox{\textwidth}{!}{%
\begin{tabular}{c l l}

\toprule
Language & Test and prediction & English gloss \\ \midrule
Russian & 
\cellbreaks{
    C: \foreignlanguage{russian}{Иногда он ходит в университет.} \\
    \foreignlanguage{russian}{Редко он ездит в театр.}\\
    Q: \foreignlanguage{russian}{Что он делает иногда?}\\
    A: \foreignlanguage{russian}{Ходит в университет.}\\~\\
    P: \foreignlanguage{russian}{Идёт в университет.}
}
& \cellbreaks{
    C: Sometimes he goes (by foot) to the university. \\
    Rarely does he go (by transportation) to the theatre.\\
    Q: What does he do sometimes? \\
    A: Goes to the university (multiple times). \\~\\
    P: Going to the university (one time).
}
\\ \midrule
Finnish & 
\cellbreaks{
    C: Äitini antoi isoäidilleni mukin. \\ 
    Isäni antoi sedälleni kameran.\\
    Q: Kenellä on uusi muki? \\
    A: Isoäidilläni. \\~\\
    P: Isoäidilleni.
}
& \cellbreaks{
    C: My mother gave my grandmother a mug. \\
    My father gave my uncle a camera.\\
    Q: Who has a new mug? \\
    A: My grandmother has. \\~\\
    P: To my grandmother.
}
\\ \midrule
Chinese & 
\cellbreaks{
    C: \begin{CJK*}{UTF8}{gbsn}桌子旁边放着六样东西，都是狗。\end{CJK*} \\
    Q: \begin{CJK*}{UTF8}{gbsn}多少狗在桌子旁边?\end{CJK*} \\
    A: \begin{CJK*}{UTF8}{gbsn}六只。\end{CJK*} \\~\\
    P: \begin{CJK*}{UTF8}{gbsn}六个。\end{CJK*}
}
& \cellbreaks{
    C: Next to the table are six things,\\
     all are dogs.\\
    Q: How many dogs are next to the table? \\
    A: Six (measure word for animals). \\~\\
    P: Six (generic measure word).
}
\\ \midrule
Swahili & 
\cellbreaks{
    C: Sadiki anakula saa nne usiku \\ 
    na anaendesha masaa matatu baadaye. \\
    Q: Anaendesha saa ngapi? \\
    A: Saa saba usiku. \\~\\
    P: Saa moja usiku.
}
& \cellbreaks{
    C: Sadiki eats at 10 PM and then drives three hours after. \\
    Q: What time does he run? \\
    A: At 1 AM. \\~\\
    P: At 7 PM.
}
\\ \midrule
\end{tabular}%
}
\caption{Examples of errors in PaLM-L predictions and English glosses for language-specific feature tests. Each example includes a context (C), question (Q), answer (A), and the model prediction (P). Tests probe motion verbs in Russian, possessives in Finnish, measure words in Chinese, and time expressions in Swahili.}
\label{tab:language-specific_feature_prediction_errors}
\end{table*}

\section{Examples of Morphological Errors} \label{app:morphological_errors}

We show example errors in predictions of PaLM-L that are due to morphology in Table \ref{tab:examples_morphological_prediction_errors_app}. 

\begin{table*}[]
\centering
\resizebox{\textwidth}{!}{%
\begin{tabular}{c l l}

\toprule
Test type, & \multirow{2}{*}{Test and prediction} & \multirow{2}{*}{English gloss}   \\ 
Language & & \\ \midrule
\cellbreaks{Negation \\ Slovak}  & 
\cellbreaks{
    C: Pavol a Oskar nie sú vedci, \\
    ale Bohuš a Miroslav sú.\\
    Q: Kto sú vedci?\\
    A: Bohuš a Miroslav. \\~\\
    P: Bohús a Miroslav.
}
& \cellbreaks{
    C: Pavol and Oskar are not scientists, \\
    but Bohuš and Miroslav are.\\
    Q: Who are scientists? \\
    A: Bohuš and Miroslav. \\~\\
    P: Bohús and Miroslav.
}
\\ \midrule
\cellbreaks{Numerals \\ Russian} & 
\cellbreaks{
    C: \foreignlanguage{russian}{На столе три груши и девять яблок.} \\
    \foreignlanguage{russian}{Елена съела одну грушу.}\\
    Q: \foreignlanguage{russian}{Сколько груш на столе?}\\
    A: \foreignlanguage{russian}{Две.}\\~\\
    P: \foreignlanguage{russian}{Два.}
}
& \cellbreaks{
    C: There are three pears and nine apples \\
    on the table. Elena ate one pear.\\
    Q: How many pears are on the table? \\
    A: Two. (Feminine Nominative) \\~\\
    P: Two. (Masculine Nominative)
} \\ \midrule
\cellbreaks{Spatial \\ Finnish} & 
\cellbreaks{
    C: Mukit ovat ikkunan päällä ja \\ 
    tietokoneet tuolin alla.\\
    Q: Missä ovat tietokoneet? \\
    A: Tuolin alla. \\~\\
    P: Tuolien alla.
}
& \cellbreaks{
    C: The mugs are on the window and the \\
    computers are under the chair.\\
    Q: Where are the computers? \\
    A: Under the chair. (Genitive Singular) \\~\\
    P: Under the chairs. (Genitive Plural)
} \\ \midrule
\cellbreaks{Temporal \\ Slovak} & 
\cellbreaks{
    C: Peter a Katarína boli vedcami, \\ 
     ale Katarína zmenila zamestnanie a teraz je kuchárkou.\\
    Q: Kým je Katarína? \\
    A: Kuchárkou. \\~\\
    P: Kuchárka.
}
& \cellbreaks{
    C: Peter and Katarína were scientists, \\
    but Katarína changed jobs and is now a cook.\\
    Q: Who is Katarína?\\
    A: Cook. (Instrumental)\\~\\
    P: Cook. (Nominative)
} \\ \midrule
\cellbreaks{Comparative \\ Finnish} & 
\cellbreaks{
    C: Jos vene olisi uudempi, Ylvä käyttäisi sitä. \\ 
    Jos pyörä olisi pienempi, Ylvä käyttäisi sitä. \\
    Q: Mitä Ylvä käyttäisi jos se olisi vähemmän vanha? \\
    A: Venettä. \\~\\
    P: Vene.
}
& \cellbreaks{
    C: If the boat was newer, Ylvä would use it.\\
    If the bike was smaller, Ylvä would use it.\\
    Q: What would Ylvä use if it was less old?\\
    A: Boat. (Partitive) \\~\\
    P: Boat. (Nominative)
} \\ 
\bottomrule
\end{tabular}%
}
\caption{Examples of morphological errors in PaLM-L predictions and English glosses for generated tests. Examples correspond to highlighted cells in Table \ref{tab:morphological_errors}. Each example includes a context (C), question (Q), answer (A), and the model prediction (P).}
\label{tab:examples_morphological_prediction_errors_app}
\end{table*}

\section{One-shot Results} \label{app:one-shot}

We show one-shot results for all models in Table \ref{tab:one-shot-model-comparison-app}. We show summary statistics of the average relative change in performance of the one-shot setting compared to the zero-shot setting for each language and model in Table \ref{tab:improvement-one-shot-vs-zero-shot}. 

\begin{table*}[]
\centering
\resizebox{\textwidth}{!}{%
\begin{tabular}{lllcccccccccccccr}
\toprule
& Test type & Model & \en & \es & \It & \fr & \de & \sv & \Fi & \sk & \ru & \zh & \sw & \ar  & Avg. & 0-shot $\Delta$ \\ \midrule
\parbox[t]{2mm}{\multirow{9}{*}{\rotatebox[origin=c]{90}{Negation}}} & \multirow{5}{*}{In context} & mT5-XXL & 99.6 & 97.3 & 98 & 97.7 & 92.1 & 98.6 & 96.6 & 97.5 & 98.3 & 73.1 & 97.8 & 63.4 & 91.9 & 35.0 \\
& & PaLM-S & 92.2 & 88.2 & 91 & 69.5 & 85.8 & 87.7 & 87.4 & 83.8 & 73.6 & 81.3 & 92.4 & 45.6 & 81.5 & 46.4 \\
 &  & PaLM-M & 99.8 & 99.9 & 99.4 & 99.9 & 99.2 & 99.5 & 99.1 & 99.1 & 96.4 & 96.9 & 88 & 61.6 & 94.9 & 26.7 \\

 & & PaLM-L & 99.7 & 100 & 99.9 & 100 & 100 & 99.8 & 100 & 99.6 & 100 & 99.7 & 99.9 & 95.1 & 99.5 & 2.9\\
 & & PaLM 2 & 99.6 & 100 & 99.9 & 100 & 100 & 99.4 & 99.9 & 98.4 & 100 & 100 & 99.9 & 98.1 & 99.6 & 1.3 \\
& \multirow{5}{*}{In question} & mT5-XXL & 75.2 & 78.4 & 74.4 & 76.9 & 74.2 & 79 & 74 & 71.6 & 77.6 & 51.7 & 75.2 & 60.3 & 72.1 & 53.2\\
& &  PaLM-S & 39.9 & 44.8 & 33.5 & 23.1 & 35.8 & 38.5 & 35.5 & 37.2 & 44.4 & 40.8 & 38.3 & 33.8 & 37.1 & 0.5 \\
 &  & PaLM-M & 78.2 & 92.6 & 93.8 & 95 & 92.6 & 96 & 94.2 & 90.9 & 73.6 & 75.1 & 81.3 & 61.8 & 85.4 & 32.5 \\
&  & PaLM-L & 97.7 & 99.8 & 99.9 & 99.4 & 99.4 & 99.3 & 99.9 & 99.7 & 99.7 & 97.8 & 98.5 & 93.6 & 98.8 & 10.4 \\
&  & PaLM 2 & 96.2 & 100 & 98.5 & 99.8 & 99.9 & 90.6 & 86.9 & 99.4 & 99.9 & 98.2 & 99.5 & 96.9 & 97.2 & 0.0 \\ \verythinrule
\parbox[t]{2mm}{\multirow{9}{*}{\rotatebox[origin=c]{90}{Numerals}}} & \multirow{5}{*}{Addition} & mT5-XXL & 8.4 & 7.1 & 0.8 & 5.5 & 2.1 & 8.5 & 7.3 & 0.6 & 10.4 & 12.4 & 1.9 & 58 & 10.4 & 5.0\\
& & PaLM-S & 20.1 & 13.3 & 13.3 & 10.7 & 21 & 22.4 & 7.1 & 4.6 & 9.2 & 10.6 & 5.9 & 9.2 & 12.3 & -11.5 \\
 &  & PaLM-M & 95.7 & 71.8 & 69.7 & 89.3 & 90.7 & 61.2 & 50.3 & 47.5 & 81.7 & 87.2 & 10.8 & 18.9 & 64.6 & 32.8 \\
 & & PaLM-L & 99.3 & 100 & 96.7 & 100 & 99.5 & 96.9 & 80.7 & 79.2 & 83.8 & 97.2 & 72.1 & 71.3 & 88.9 & 11.2 \\
 & & PaLM 2 & 100 & 100 & 100 & 100 & 100 & 100 & 99.9 & 98.3 & 99.8 & 100 & 89.6 & 95.1 & 98.4 & 7.3 \\
& \multirow{5}{*}{Subtraction} & mT5-XXL & 29.6 & 27 & 8.2 & 26.7 & 22.7 & 25.5 & 20.7 & 0.6 & 9.1 & 19.4 & 1.3 & 42.1 & 18.5 & -4.5 \\
& & PaLM-S & 25 & 23.4 & 18.6 & 23.9 & 25.4 & 21.7 & 13.8 & 16.3 & 16.7 & 16.1 & 10.8 & 14.5 & 18.9 & 2.6 \\
 &  & PaLM-M & 56.7 & 64.1 & 61.3 & 58.8 & 49.7 & 29.8 & 34.7 & 26.7 & 40.6 & 38.8 & 11.7 & 38.2 & 42.6 & 6.2 \\
&  & PaLM-L & 92.5 & 94.3 & 94.8 & 97 & 79.2 & 96.2 & 95.4 & 85.8 & 86.4 & 89.9 & 39.2 & 64.2 & 83.9 & 7.5\\ 
&  & PaLM 2 & 99.9 & 99.8 & 100 & 99.9 & 96.1 & 100 & 98.6 & 99.8 & 99 & 88.1 & 60.9 & 98.9 & 94.6 & 6.9 \\ \verythinrule
\parbox[t]{2mm}{\multirow{9}{*}{\rotatebox[origin=c]{90}{Spatial}}} & \multirow{5}{*}{Prepositions} & mT5-XXL & 91.1 & 90.6 & 66.8 & 93.4 & 22.7 & 84.7 & 73.8 & 2.6 & 41 & 85.8 & 9.1 & 81.2 & 59.2 & 3.5 \\
& & PaLM-S & 79.9 & 50 & 52.8 & 54.6 & 67.8 & 48.5 & 52 & 45.8 & 53.7 & 97 & 42.4 & 41.2 & 57.1 & -1.0 \\
 &  & PaLM-M & 99.6 & 97 & 97.5 & 99.2 & 94.6 & 92.5 & 83.7 & 93.8 & 93.7 & 97.9 & 77.2 & 78.1 & 92.1 & 9.6 \\
&  & PaLM-L & 100 & 100 & 100 & 100 & 100 & 100 & 99.2 & 100 & 100 & 100 & 99.3 & 97.7 & 99.7 & 2.6\\
& & PaLM 2 & 100 & 100 & 100 & 98.4 & 100 & 100 & 100 & 100 & 100 & 100 & 100 & 100 & 99.9 & 0.5\\
& \multirow{5}{*}{Position} & mT5-XXL & 98.1 & 100 & 95.8 & 99.9 & 98.5 & 97.4 & 99.9 & 100 & 100 & 96.1 & 72.6 & 73.1 & 93.9 & 36.3\\
& & PaLM-S & 85.5 & 67.2 & 66.4 & 59.1 & 68.7 & 66.7 & 88 & 75 & 38.3 & 99 & 53.6 & 36.7 & 67.0 & 27.4 \\
 &  & PaLM-M & 99.9 & 93.5 & 99.6 & 100 & 99.6 & 91 & 98.6 & 98.7 & 93.1 & 99.6 & 97 & 81.6 & 96.0 & 46.4 \\
& & PaLM-L & 100 & 99.9 & 99.9 & 99.9 & 100 & 99.9 & 99.8 & 100 & 99.5 & 99.9 & 85.9 & 81.2 & 96.9 & 31.4\\
& & PaLM 2 & 100 & 100 & 100 & 99.9 & 100 & 100 & 100 & 100 & 100 & 100 & 99.9 & 99.9 & 100 & 5.7\\ \verythinrule
\parbox[t]{2mm}{\multirow{9}{*}{\rotatebox[origin=c]{90}{Temporal}}} & \multirow{5}{*}{Past} & mT5-XXL & 90.4 & 99.1 & 90.9 & 96.4 & 93.8 & 87.7 & 87 & 97.7 & 91.5 & 90.9 & 86.5 & 75.7 & 90.7 & 28.8 \\
& & PaLM-S & 95.1 & 75.1 & 84.2 & 94.4 & 44.6 & 84 & 57.7 & 32.6 & 96.8 & 78 & 77.7 & 54.9 & 72.9 & 20.3 \\
 &  & PaLM-M & 94.5 & 91.8 & 61.7 & 75 & 79.7 & 53.2 & 41.5 & 60.5 & 34.3 & 84.5 & 84.4 & 74.8 & 69.7 & -18.7 \\
& & PaLM-L & 99.8 & 98.6 & 97.2 & 97 & 99.6 & 99.8 & 99.9 & 96.1 & 99 & 99.9 & 100 & 99.4 & 98.8 & 2.7\\
& & PaLM 2 & 100 & 99.9 & 100 & 100 & 100 & 100 & 100 & 100 & 99.9 & 100 & 100 & 99.4 & 99.9 & 7.9\\
& \multirow{5}{*}{Future} & mT5-XXL & 90.7 & 96.4 & 98 & 92.7 & 93.4 & 91 & 84.5 & 91.3 & 89.5 & 86.5 & 89.6 & 61.2 & 88.6 & 30.4\\
& & PaLM-S & 98.6 & 54.9 & 57.4 & 90.3 & 45 & 71.1 & 44.7 & 13.5 & 98.7 & 72.8 & 79.2 & 59.3 & 65.5 & 5.9 \\
 &  & PaLM-M & 92.4 & 93.5 & 93.2 & 69.8 & 90.2 & 58.4 & 55.5 & 61.4 & 47.8 & 62.1 & 42.2 & 78.2 & 70.4 & -12.0 \\
& & PaLM-L & 100 & 97.4 & 98.8 & 92.5 & 99.3 & 100 & 100 & 93.2 & 99.2 & 99.7 & 100 & 98.4 & 98.0 & 2.5\\
& & PaLM 2 & 100 & 100 & 100 & 100 & 100 & 100 & 100 & 100 & 100 & 100 & 100 & 100 & 100 & 0.6\\ \verythinrule
\parbox[t]{2mm}{\multirow{9}{*}{\rotatebox[origin=c]{90}{Comparative}}} & \multirow{5}{*}{Standard} & mT5-XXL & 86 & 90.3 & 92.3 & 81.7 & 82.7 & 83.7 & 92.3 & 85 & 89.3 & 85.3 & 81 & 71.2 & 85.0 & 47.1\\
& & PaLM-S & 90.3 & 97.7 & 90 & 87.3 & 94 & 77 & 76 & 89.3 & 89.3 & 56.3 & 98 & 47.2 & 82.7 & 20.4 \\
 &  & PaLM-M & 88 & 90.3 & 97.7 & 84 & 91 & 93.3 & 80.7 & 83 & 91.7 & 88.7 & 97 & 87.1 & 89.4 & 19.0 \\
& & PaLM-L & 100 & 100 & 100 & 99.7 & 100 & 100 & 100 & 99.7 & 100 & 99.7 & 100 & 94.9 & 99.5 & 3.0 \\
& & PaLM 2 & 100 & 100 & 100 & 100 & 100 & 99.3 & 100 & 100 & 100 & 100 & 100 & 100 & 99.9 & 0.2\\
& \multirow{5}{*}{Conditional} & mT5-XXL & 47.4 & 72.2 & 73.9 & 64.5 & 61.7 & 74.4 & 72.7 & 62.6 & 42.4 & 69.1 & 73.5 & 74.5 & 66.7 & 43.2\\
& & PaLM-S & 86.2 & 74.9 & 60.5 & 70.1 & 38.5 & 70.6 & 88.6 & 43.8 & 36.7 & 78.5 & 77.2 & 55.3 & 65.1 & 44.7 \\
 &  & PaLM-M & 72.5 & 82.9 & 76 & 67.3 & 76.3 & 60.1 & 61.3 & 52.6 & 39.9 & 73.8 & 78.6 & 71.3 & 67.7 & 22.8 \\
& & PaLM-L & 75.7 & 70.4 & 83.2 & 70.8 & 53.1 & 47.2 & 48.9 & 35.6 & 30.6 & 63 & 50.3 & 82.8 & 57.8 & 8.0 \\
& & PaLM 2 & 93.7 & 96.1 & 97.2 & 93.9 & 83.5 & 82.3 & 86.9 & 22.5 & 47.8 & 93.1 & 77.4 & 85.1 & 78.7 & 14.6 \\
 \bottomrule
\end{tabular}%
}
\caption{Accuracy (in \%) of mT5-XXL, PaLM-S, PaLM-M, PaLM-L, and PaLM 2 on generated tests in a one-shot setting. The right-most column shows the relative change compared to the zero-shot setting for each model.}
\label{tab:one-shot-model-comparison-app}
\end{table*}

\begin{table*}[]
\centering
\resizebox{\textwidth}{!}{%
\begin{tabular}{llllllllllllll}
\toprule
 & \en & \es & \It & \fr & \de & \sv & \Fi & \sk & \ru & \zh & \sw & \ar & Avg \\ \midrule
mT5-XXL & 20.3\% & 136.8\% & 59.2\% & 77.5\% & 27.8\% & 85.9\% & 58.1\% & 114.2\% & 66.0\% & 67.7\% & 92.1\% & 22.1\% & 69.0\% \\
PaLM-S & 7.3\% & 51.7\% & 55.3\% & 21.8\% & 11.8\% & 10.4\% & 38.5\% & 85.2\% & 64.5\% & 40.9\% & 145.6\% & 35.1\% & 47.3\% \\
PaLM-M & 3.9\% & 23.8\% & 41.3\% & 7.2\% & 20.2\% & 11.0\% & 30.9\% & 41.3\% & 28.2\% & 46.0\% & 90.4\% & 33.4\% & 31.5\% \\
PaLM-L & 4.3\% & 7.3\% & 8.8\% & 3.9\% & 7.3\% & 3.6\% & 5.7\% & 15.8\% & 11.6\% & 15.5\% & 19.7\% & 12.5\% & 9.7\% \\
PaLM 2 & 0.9\% & 1.4\% & 6.4\% & 0.9\% & 3.1\% & 0.2\% & 9.7\% & 3.8\% & 1.7\% & 10.9\% & 10.6\% & 6.7\% & 4.7\% \\
\bottomrule
\end{tabular}%
}
\caption{Average relative improvement of the one-shot vs the zero-shot setting for all models across all languages.}
\label{tab:improvement-one-shot-vs-zero-shot}
\end{table*}

\end{document}